\documentclass[conference]{IEEEtran}
\IEEEoverridecommandlockouts
\usepackage{amsmath,amssymb,amsfonts}
\usepackage{algorithmic}
\usepackage{graphicx}
\usepackage{textcomp}
\usepackage{xcolor}
\usepackage{float}
\usepackage{tcolorbox}
\usepackage{tabularx}
\usepackage{fancyhdr}
\makeatletter

\def\ps@IEEEtitlepagestyle{
  \def\@oddfoot{\mycopyrightnotice}
  \def\@evenfoot{}
}
\def\mycopyrightnotice{
  {\footnotesize CICT2023-78~\copyright~IEEE\hfill} 
  \gdef\mycopyrightnotice{}
}
\@ifundefined{showcaptionsetup}{}{
 \PassOptionsToPackage{caption=false}{subfig}}
\usepackage{subfig}
\makeatother

\usepackage{eso-pic}
\newcommand\AtPageUpperMyright[1]{\AtPageUpperLeft{
 \put(\LenToUnit{0.5\paperwidth},\LenToUnit{-1cm}){
     \parbox{0.5\textwidth}{\raggedleft\fontsize{9}{11}\selectfont #1}}
 }}
\newcommand{\conf}[1]{
\AddToShipoutPictureBG*{
\AtPageUpperMyright{#1}
}
}
\usepackage[style=ieee]{biblatex} 
\def\BibTeX{{\rm B\kern-.05em{\sc i\kern-.025em b}\kern-.08em
    T\kern-.1667em\lower.7ex\hbox{E}\kern-.125emX}}
\begin{document}

\title{RoadScan: A Novel and Robust Transfer Learning Framework for Autonomous Pothole Detection in Roads
}
\conf{IEEE 7th Conference on Communication and Information Technology}

\author{\IEEEauthorblockN{Guruprasad Parasnis, Anmol Chokshi, Vansh Jain, Kailas Devadkar}
\IEEEauthorblockA{Sardar Patel Institute of Technology, Mumbai, India}
}
\maketitle

\begin{abstract}
In the past few years, there have been numerous accidents due to potholes on the road. Many methods have been attempted to remove these potholes, but these methods are time-consuming. Hence, a driver should detect potholes from a safe distance in order to avoid damage. Existing methods for pothole detection heavily rely on object detection algorithms which tend to have a high chance of failure owing to the similarity in structures and textures of a road and a pothole. Additionally, these systems utilize millions of parameters thereby making the model difficult to use in small-scale applications for the general citizen. This research paper presents a novel approach to pothole detection using deep learning and image processing techniques. The proposed system leverages the VGG16 model for feature extraction and utilizes a custom Siamese network with triplet loss, referred to as RoadScan. Evaluation metrics such as accuracy, EER, precision, recall, and AUROC validate the effectiveness of the system. Additionally, the proposed model demonstrates computational efficiency and cost-effectiveness by utilizing fewer parameters and data for training. The network proposed in this model performs with a 96.12\% accuracy, 3.89\% EER, and a 0.988 AUROC value, which is highly competitive with other state-of-the-art works.
\end{abstract}


\section{Introduction}
In recent years, technology has played a crucial role in advancing automated systems across different sectors. The introduction of autonomous systems has significantly improved the convenience and efficiency of human lives. Among various sectors benefiting from automation, transportation and surveillance systems have witnessed remarkable progress. Within the realm of transportation, roads hold a vital position as they form an extensive network. Ensuring the safety of users is a top priority for autonomous systems, and one of the significant threats on roads is the presence of potholes.

According to official data released by the Government of India, in 2022 alone, 1481 accidents resulted in loss of life, and 3103 individuals were injured due to potholes. These statistics underscore the critical importance of road maintenance. However, the COVID-19 pandemic and subsequent lockdowns have adversely affected several sectors, including road maintenance. Consequently, road conditions have deteriorated, necessitating the development of an autonomous system capable of monitoring and detecting road conditions.
This paper proposes a pothole detection system that leverages deep learning and image processing techniques. In recent times, numerous deep learning-based object detection methods have emerged, utilizing Convolutional Neural Networks (CNNs) for feature extraction. Various versions of pre-trained algorithms such as YOLO, DenseNet, and RetinaNet have been trained on diverse datasets encompassing both water-logged and dry potholes of varying shapes and sizes. The performance of these models is evaluated using accuracy, precision, recall, AUROC, and other relevant metrics. The models exhibit reasonable accuracy in detecting a wide range of potholes. This paper proposes a novel transfer learning approach to the detection of potholes with features extracted from the VGG16 model and passed onto a unique, custom Siamese network architecture that is efficient and economical and gives a state-of-the-art performance. 
Overall, this research highlights the significance of technology in the development of automated systems, particularly in the transportation sector. By employing advanced deep learning and image processing techniques, the proposed pothole detection system aims to mitigate the risks associated with potholes, enhance road safety, and contribute to the overall convenience and well-being of individuals utilizing road transportation.

\section{Related Work}
Various deep learning algorithms have been developed for the implementation of pothole detection. Due to a shortage of real pothole data, various augmentation techniques in combination with a deep CNN ~\cite{1} have been proposed, giving a strong 98\% accuracy. The most famous pothole detection algorithm is the YOLO (with its various versions like YOLOv3, YOLOv4, etc). ~\cite{2}, ~\cite{9}, ~\cite{12}, ~\cite{19} are some of the instances where YOLO is used to detect potholes. ~\cite{2} gives accuracy in the range of 33\% to 69\%, ~\cite{9} gives an f1-score of 0.51, ~\cite{12} discusses results obtained with a modified YOLOv2 architecture, achieving precisions and recalls of 0.89 and 0.87 respectively with 35 million parameters, signifying that although YOLO is a good algorithm for pothole detection, it largely depends on the data available and how that data is used. In cases where it does perform very well, the number of parameters required is huge, which tends to reduce the economy of the model. ~\cite{3} proposes pothole detection with a DenseNet121 architecture, achieving 99.3\% accuracy, but significantly increasing the number of parameters involved in the model. An example of how sensory data can be integrated with a CNN is given by ~\cite{4} but does not give promising results with an 87.20\% precision, 92.7\% recall, and 89.9\% F1-Score. Some works also include detecting areas of interest in the pothole dataset proceeding by training a CNN architecture ~\cite{5}. Experimental results in this work give an overall accuracy of 90\% for the computed area of the pothole when compared with the actual measured area with ± 10\% deviation. Works that have various types of road data along with present potholes ~\cite{7} and use a CNN also have been tested with a high (96\%) accuracy in detecting the type of road but do not repeat this good performance in detecting potholes, with an accuracy of 88\%. ~\cite{10} is a notable work that proposes a lightweight model for the implementation of this task, but gives an accuracy of only 86.29\%, which is a compromise on the efficiency for the purpose of reducing the number of parameters in the network. The U-Net architecture proposed in ~\cite{13} gives a very good result with an accuracy of 97\% which illustrates the importance of deep learning networks in alleviating this task of detecting potholes on the roads. There has also been an attempt at simulating the detection of potholes in a virtual environment ~\cite{15} which achieves stupendous results, obtaining a (99.80 \%) of Accuracy, Precision (100\%), Recall (99.60\%), and F-Measure (99.60\%) simultaneously. But, as these results are simulated in a controlled environment, they cannot be emulated in real-world applications directly. Finally, ~\cite{18} gives another example of using some image processing techniques, using the Single Shot Detector with a feed-forward neural network, bagging a commendable 96.7\% accuracy. A proper prototype model has been developed by Dewangan et al. \cite{21}. The prototype is developed using a CNN with a vision camera to explore and validate the potential and autonomy of its driving behavior in the prepared road environment. The experimental analysis of the proposed model on various performance measures have obtained accuracy, sensitivity, and F-measure of 99.02\%, 99.03\%, and 98.33\%, respectively, which are comparable with the available state-of-art techniques. A layered approach in the CNN enables the architecture to create a feature map that is updated in every combination of convolutional pooling and fully connected layers. These feature maps represent the characteristics of intended objects. Another major breakthrough was achieved by Dewangan et al. \cite{22} where they designed a novel two-tier deep learning-based lane detection framework for images at different weather conditions. In both tiers, the Local Vector Pattern (LVP) based texture features are extracted and an Optimized Deep Convolutional Neural Network (DCNN) is utilized to classify roads and lanes as well. Dewangan et al. \cite{23} lead another major research in pothole detection where they exploit a siamese fully convolutional network based on VGG-net architecture that considers semantic contour, RGB channel and location prior to segmenting road regions precisely. As a major contribution, superpixel segmentation was carried out, where the RGB images are given as input to the FCN network and the road regions of images are set as a target. Further, the segmented outputs are fused using AND operation to attain the final segmented output that detects the road regions accurately. The presented model has achieved a minimal value of negative measures and accuracy is 8.2\% higher than traditional methods.

\section{Proposed Methodlogy}

\subsection{Dataset}\label{AA}
The dataset used in this paper has been obtained from Kaggle. It contains two classes: normal and potholes. The normal instances have 352 images and the pothole instances have 329 images. The paper uses 280 potholes and 280 normal images for training and validation. The remaining 72 for normal and 49 for potholes are used for testing. 

\subsection{Data Pre-Processing}
The data preprocessing involves resizing the images to a fixed shape $(224,224,3)$ due to irregularities in the dataset. As the images are of varied settings, taken under various conditions, the paper employs some image processing techniques like thresholding with the help of Otsu's method. 
Fig. 1 showcases the output of Otsu's thresholding on a sample image of the dataset which contains a pothole. 
\begin{figure}[htbp]
\centering
\includegraphics[width=0.5\textwidth,height = 5 cm]{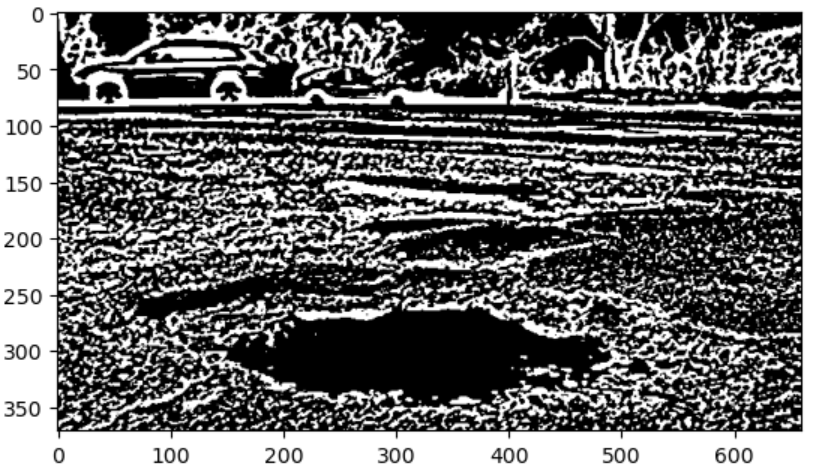}
\caption{Otsu's Thresholding on a sample image}
\label{fig1}
\end{figure}
Otsu's thresholding method is a popular image processing technique for converting grayscale images to binary images by converting pixel values above a certain threshold to white and those below it to black. Adaptive thresholding is a type of thresholding in which the threshold value for each pixel is adjusted based on the local intensity of its surroundings. This technique is useful for our images with varying diversity in the roads, objects and potholes. \par 

\begin{figure*}
\centering
  \includegraphics[width=1\linewidth]{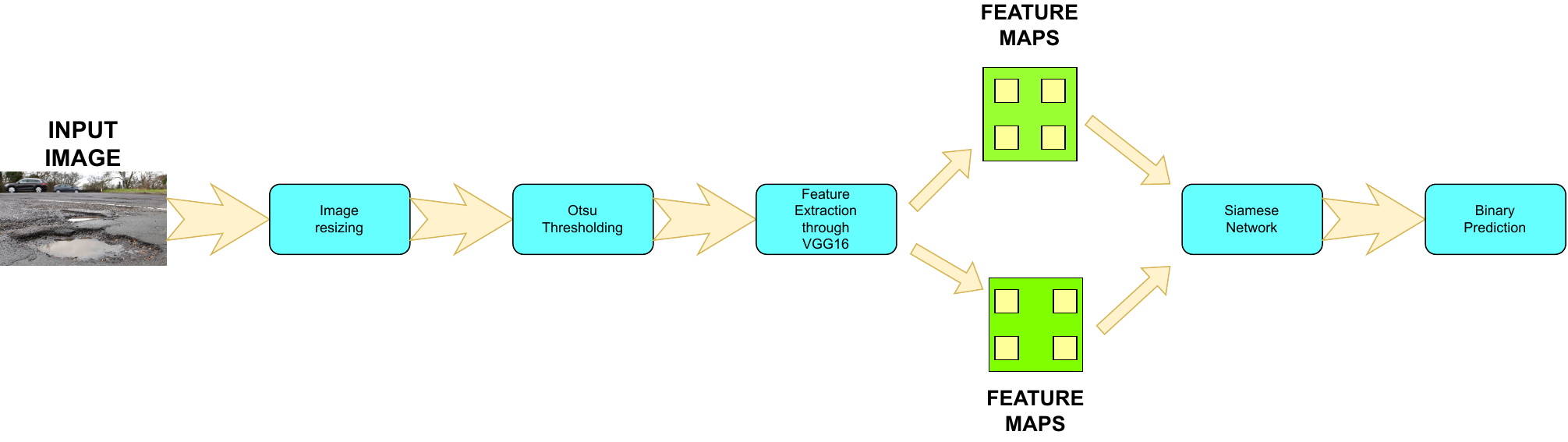}
  \caption{Process Flow Diagram}
  \label{process_diagram}
\end{figure*}

\subsection{Feature Extraction}
As Karen et al. ~\cite{20} signifies, VGG16 is a useful model for the extraction of features from an image and using these features to further employ in a different model. This paper uses the VGG16 model to extract the textural features, especially of the pothole, since a pothole is predominantly a texture-based image. One of the reasons for resizing the input images to a fixed size of ($224 \times 224$) is due to the fact that a VGG16 model works perfectly with that shape. The feature extraction process occurs at the later convolutional layers of VGG16, which capture higher-level and more abstract features. These layers are deeper in the network and have larger receptive fields. The earlier layers capture low-level features like edges and textures, while the deeper layers capture high-level features like shapes and objects. At each convolutional layer, the VGG16 model produces a set of feature maps, also known as activation maps. These feature maps represent the activation of each filter in the layer across the spatial dimensions of the input image. They highlight regions in the pothole image that are particularly relevant to the learned features. To obtain a fixed-length feature vector for the entire image, a global average pooling operation is typically applied to the last convolutional layer's feature maps. This operation averages each feature map spatially, resulting in a reduced-dimensional representation. The output of the last global average pooling layer can be considered as the extracted features from the VGG16 model. These features capture high-level information about the input pothole image, which is then used for further fine-tuning the pothole dataset.

\subsection{Proposed Deep Learning Network}
This paper proposes a novel transfer learning network using a Siamese model. Fig. 2 provides the complete process flow diagram of the approach in this paper. As described in the Feature Extraction section, the features are extracted from the input images using the VGG16 pre-trained model, which are in turn fed to a Siamese network. The Siamese model is an image comparison model that takes two images as input and outputs a similarity score between them. The Siamese network architecture is made up of two identical sub-networks with the same weights. The input images are fed into each sub-network, and the resulting feature vectors are compared using a distance metric. The network is trained with triplet loss, which reduces the distance between similar images while increasing the distance between dissimilar images. \par 
Two functions are defined as same-pair and diff-pairs to train our neural network. The same-pair function creates the same pairs of images by selecting two images at random from a batch of similar images, with no repeated pairs allowed. The diff-pairs function generates unique image pairs by randomly selecting two images from different batches, with no repeated pairs permitted. The same pairs are labeled with a 1 and different pairs with a 0. The model is fed with all possible same and different pairs which makes a combined 100000 pairs for training. \par 
Following that, the extracted features are passed through a series of convolutional layers that include batch normalization and max-pooling layers. To prevent overfitting, the output of these layers is flattened and passed through two fully connected layers with dropout regularization. 
To avoid overfitting, the Siamese model is trained using triplet loss with a margin of one and an early stopping technique. For the optimizer, the RMSprop optimizer was employed. The model recognized similar textures in images containing potholes with high accuracy. Based on the output given by this model, the pothole images are classified as “positive” or “negative”. The model has also been trained using different types of losses like the contrastive loss to get a better insight into the results of the network.
For one step of the Siamese training, a total of three images are fed
in. These images are: anchor image (A); positive class image
(P), which belongs to the same person as that of anchor image;
negative class image (N), which belongs to a person other than
that of the anchor image. The loss function used in the proposed model is given in equation (1). Fig. 3 showcases how the images are fed into the proposed model. \par 
\begin{equation}
   (Y, D) = \frac{1}{2N} \sum_{i=1}^{N} Y \cdot D^2 + (1 - Y)\cdot \max(margin - D, 0)^2 
\end{equation} \par 
These images are then accordingly fed into the loss function to evaluate how close the input images are to the anchor image. This novel proposed network has given efficient results.  \par 

\begin{figure}[htbp]
\centering
\includegraphics[width=0.5\textwidth]{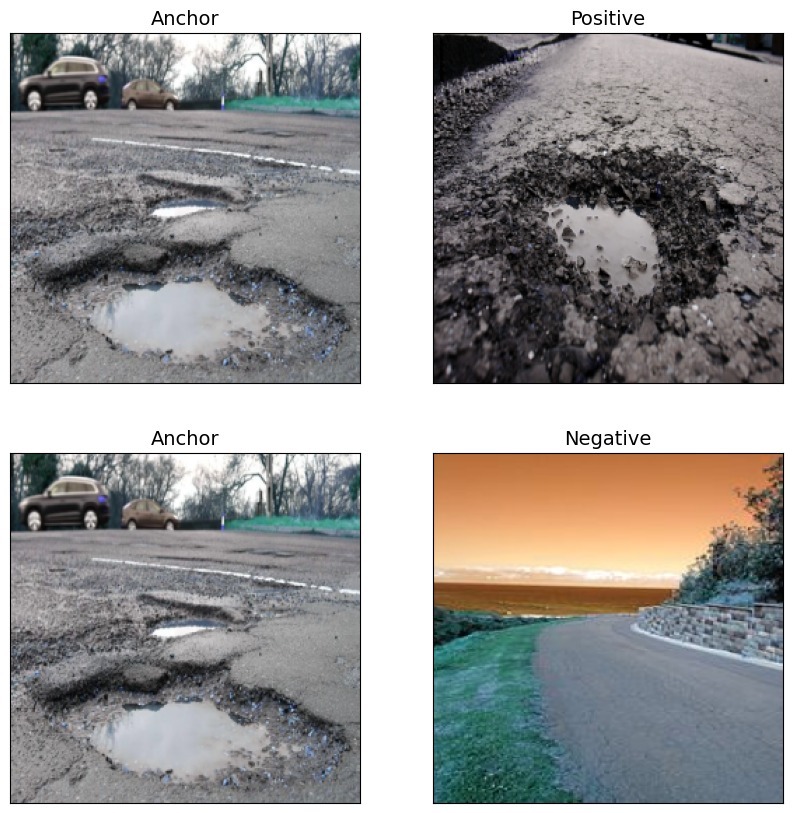}
\caption{Input through triplet loss}
\label{fig}
\end{figure} \par 
\begin{figure*}
\centering
  \includegraphics[width=1\linewidth]{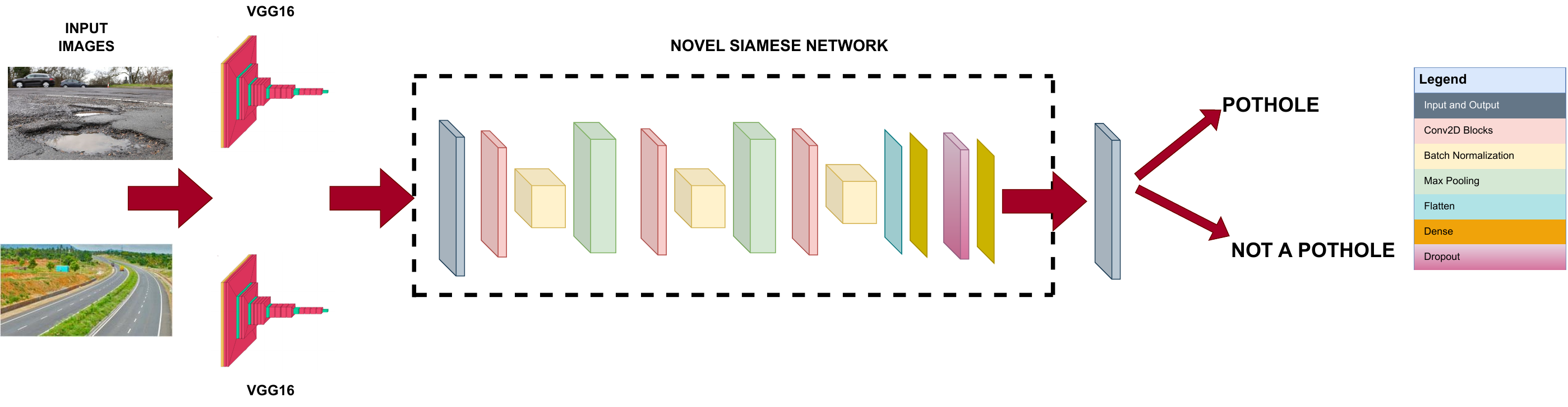}
  \caption{Proposed Deep learning network}
  \label{architecture}
\end{figure*} 
These are sample images from the dataset that are fed for training into the model. \par
The network starts with taking 2 input images owing to the Siamese network. Each image is fed into the first convolution layer, which has a kernel size of ($3 \times 3$), the number of filters as 16, and a stride of 1. The output is then batch normalized and fed into max pooling to avoid overfitting. All the max pooling layers have a pool size of ($2 \times 2$) and stride as 2. This max pooled output layer is then sent to another convolution layer with the number of filters as 32, kernel size as ($3 \times 3$) and stride as 1. We again perform the batch normalization and max pooling blocks. We repeat the entire process of applying the convolution layer, batch normalization, and max pooling layers. This time the convolution layer had a number of filters as 64, a kernel size of ($3 \times 3$) and a stride of 1. Finally, we flatten the output, transposing the input matrix into a single channel configuration and applying dropout (having a rate of 0.5) and dense layers. The final output gives a probability as to whether the potholes are present or not. Fig. 4 shows the entire schematic of the proposed architecture.


\subsection{Performance Metrics}
This section aims to explain the metrics used for evaluating the performance of the proposed pothole detection system in this work. The experimental approach involves modifying the network while keeping the other components of the pipeline unchanged. To assess the performance, well-established metrics such as Equal Error Rate (EER), Area Under Receiver Operating Characteristics (AUROC), Area Under Precision-Recall (AUPR), accuracy, precision, recall, and the F1-Score are employed.

By utilizing these metrics, the experimental evaluations focus exclusively on the network's performance and efficiency while keeping the remaining pipeline components unchanged. The analysis of these metrics not only quantifies the network's performance but also assists in selecting and optimizing the most suitable network for the pothole detection pipeline.

In this research, careful consideration is given to the number of actual genuine pairs and imposter pairs that can be formed within the context of the study. The total number of actual genuine pairs is determined by multiplying the number of training images, validation images, and pothole images in the dataset (referred to as $UVP$). Similarly, the number of actual imposter pairs is calculated by subtracting the number of actual genuine pairs from the total number of pairs, which is $UVP^{2} - UVP$. Importantly, the ground truth information regarding whether these pairs belong to the same class or not is known.

To determine whether a pair is considered to belong to the same class, the detection system compares the distance between their respective features to a predetermined threshold value denoted as $T$. By sweeping across various values of $T$, the precision-recall curve can be constructed, providing insights into the system's performance across different threshold settings.

To quantify the performance of the pothole detection system, two widely recognized metrics are employed: the Area Under Precision-Recall (AUPR) curve and the Equal Error Rate (EER). AUPR measures the overall performance of the system by calculating the area under the precision-recall curve, which reflects the trade-off between precision and recall. On the other hand, EER represents the point at which the False Acceptance Rate (FAR) is equal to the False Rejection Rate (FRR), serving as a pivotal point for evaluating system performance.

By utilizing these metrics, the detection performance of the pothole detection system can be effectively evaluated and analyzed. EER, AUROC, AUPR, precision, recall, and F1-Score provide valuable insights into the system's ability to accurately identify genuine pairs and distinguish them from imposter pairs. These evaluation measures contribute to a comprehensive assessment of the system's performance and assist in identifying the optimal threshold setting for achieving desired verification outcomes.

To summarize, this research paper employs AUPR and EER values as quantitative metrics for evaluating the detection and classification performance of the proposed pothole detection system, while taking into account the modifications made to the network.
\begin{table*}[t]
\centering
\resizebox{\linewidth}{!}{
\begin{tabular}{|c|c|c|c|}
\hline
\textbf{Previous networks} & \textbf{Dataset} & \textbf{Algorithms} & \textbf{Results} \\
\hline
Manalo et al. \cite{2} & Pothole dataset from Kaggle & Transfer Learning using YOLOv2 & 95.3\% average precision with 33\% to 69\% accuracy \\
\hline
Arjapure et al. \cite{5} & Pothole dataset from Kaggle & Region of Interest Extraction + CNN & 92\% accuracy \\
\hline
Pereira et al. \cite{13} & Custom database collected from ASUS\_z01RD phone & U-Net architecture & 97\% accuracy \\
\hline
Ping et al. \cite{14} & Custom database & Single Shot Detector + YOLO & 82\% accuracy \\
\hline
Rastogi et al. \cite{16} & Specially curated database from an Indian city Bangalore & Modified YOLOv2 & 0.87 precision and 0.89 recall \\
\hline
Agrawal et al. \cite{7} & Kaggle Pothole database & CNN based approach & 88\% accuracy in detecting potholes  \\
\hline
Karen et al. \cite{20} & Custom curated database & Faster R-CNN using VGG16 & 0.79 precision and 0.80 recall \\
\hline
\textbf{Proposed Network} & \textbf{Kaggle Pothole database} & \textbf{Novel VGG16 + Siamese network transfer learning} & \textbf{3.89\% EER, 96.1\% accuracy, 0.985 AUPR, 0.988 AUROC} \\
\hline
\end{tabular}
}
\vspace{0.3 em}
\caption{Comparison table for previous research based on dataset, processing algorithms, and results}
\label{Relatedwork_metrics}
\end{table*}

\begin{table*}[t]
\centering
\resizebox{\linewidth}{!}{
\begin{tabular}{|c|c|c|c|c|c|c|c|c|}
\hline
\textbf{Network used} & \textbf{Trainable Parameters} & \textbf{AUROC} & \textbf{AUPR} & \textbf{EER} & \textbf{Accuracy} & \textbf{Precision} & \textbf{Recall} & \textbf{F1-Score} \\
\hline
VGG16 + Siamese Network with a kernel regularizer & 265,120 & 0.981 & 0.981 & 6.04\% & 93.7\% & 0.939 & 0.936 & 0.937 \\
\hline
VGG16 + Siamese Network with no kernel regularizer & 265,132 & 0.987 & 0.983 & 4.29\% & 95.5\% & 0.957 & 0.953 & 0.955 \\
\hline
VGG16 + Siamese Network with no Dropout and only 1st block of CNN & 361,248 & 0.989 & 0.986 & 4.52\% & 95.1\% & 0.954 & 0.946 & 0.95 \\
\hline
VGG16 + Siamese Network with only 1 CNN block and Dropout in the last layer & 361,248 & 0.98 & 0.973 & 5.16\% & 94.7\% & 0.948 & 0.946 & 0.947 \\
\hline
VGG16 + Siamese Network with 3 CNN blocks and Dropout in the last layer & 196,272 & 0.972 & 0.964 & 5.67\% & 94.7\% & 0.948 & 0.946 & 0.947 \\
\hline
VGG16 + Siamese Network with Dropout in all layers & 161,366 & 0.974 & 0.962 & 7.5\% & 92.4\% & 0.925 & 0.924 & 0.924  \\
\hline
VGG16 + Siamese Network with only Flatten and Dense Layers & 64,55,680 & 0.973 & 0.958 & 7.29\% & 92.7\% & 0.927 & 0.927 & 0.927 \\
\hline
VGG16 + Siamese Network with no kernel regularizer and 2x filters & 64,55,680 & 0.983 & 0.977 & 5.91\% & 94.1\% & 0.941 & 0.941 & 0.941 \\
\hline 
VGG16 + Siamese Network with no kernel regularizer and 2 extra blocks & 734,146 & 0.977 & 0.965 & 5.93\% & 94.08\% & 0.941 & 0.941 & 0.941 \\
\hline 
\textbf{Proposed Network (RoadScan)} & \textbf{372,448} & \textbf{0.988} & \textbf{0.985} & \textbf{3.89\%} & \textbf{96.12\%} & \textbf{0.961} & \textbf{0.961} & \textbf{0.961} \\
\hline
\end{tabular}
}
\vspace{0.3 em}
\caption{Comparison table for final model selection}
\label{performance_metrics}
\end{table*}

\subsection{Experimental Results and Tables}
The proposed model was evaluated against previous state-of-the-art works and it outperformed all the networks and results in terms of the performance metrics. The proposed network performs with an unmatched 3.89\% EER, 96.12\% testing accuracy, 0.985 AUPR and 0.988 AUROC values. The network proposed by the paper, which includes a transfer learning technique of extracting features from images using VGG16 and feeding those features to a Siamese network is a novel architecture that gives a competitive performance surpassing state of the art. Table I demonstrates the proposed network put up against previously researched works and their performances. The model proposed in the paper outweighs the other comparative networks by huge margins. Fig. 5 provides a graphical representation of the obtained results. \par 
\begin{figure}[htbp]
\centering
\includegraphics[width=0.5\textwidth]{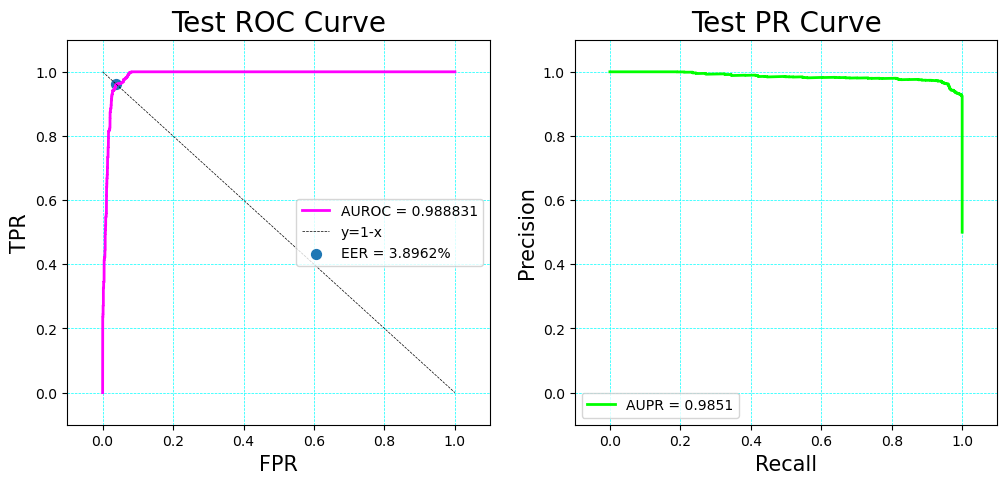}
\caption{AUROC, AUPR and EER of the proposed network}
\label{fig2}
\end{figure}

\subsection{Economy and Efficiency Analysis of the System}
This paper proposes not only an efficiently performing network but also a state-of-the-art economic network. The proposed network is economical in the sense that it contains very few trainable parameters and still provides excellent efficiency. Table II provides a detailed analysis of how the network is tested in various conditions and how the pruning of the network leads to lesser parameters and better performance, ultimately leading to an optimal network that is the most efficient while maintaining a good economy. The paper demonstrates different networks tried and tested to maintain a robust system for pothole detection. 7 different performance metrics have been obtained, which is not done by any of the cited works. All the values provide evidence of the superior performance of the network. The proposed and optimal approach gives a testing accuracy of 96.12\%. The analysis demonstrates the optimal nature of the network that is introduced in the paper. Not only does it give as less as 372,448 parameters, it gives top-notch and efficacious results to solidify its stature as a robust state-of-the-art system. In contrast, heavy networks like YOLOv2 as seen from \cite{2}, \cite{12} and \cite{19} have lakhs of parameters and still fail to provide a better result than the proposed network. Networks that do perform very well in terms of performance metrics like accuracy and precision \cite{13} do not take into account the economy of the model. In this paper, it has been experimentally demonstrated that the proposed model not only maintains its economy by drastically reducing the number of trainable parameters but also doesn't compromise its efficiency in performance.

\section{Conclusion} 
In this paper, a novel network using a transfer learning approach: VGG16 to extract features and a Siamese network with triplet loss, which we call RoadScan, has been proposed. The proposed network performs with an unmatched 3.89\% EER and 96.12\% accuracy on the dataset used. The proposed model is computationally efficient as the results obtained are comparable and better than the state-of-the-art with the added advantage that it is economical as it only utilizes much lesser parameters, and lesser data to train upon. 
Through our experiments, the capability of our network to learn a custom generalized, lightweight and economical model for pothole images is showcased. Using this approach, our method has been able to successfully classify potholes and normal roads and the obtained detection performance is better than that of non-processed images and simple CNNs. As a part of future work, we will explore the utility of other types of networks and other road obstacles such as speedbumps.

\printbibliography

\end{document}